\pdfoutput=1

\documentclass[11pt]{article}

\usepackage[]{acl}

\usepackage{times}
\usepackage{latexsym}

\usepackage[T1]{fontenc}

\usepackage[utf8]{inputenc}

\usepackage{color}
\usepackage{tabularx}
\usepackage{tcolorbox}
\usepackage{xcolor}
\usepackage{colortbl}
\usepackage{tabu}
\usepackage{booktabs}
\usepackage{multirow}

\usepackage{graphicx} 
\usepackage{float}
\usepackage{subfigure} 
\usepackage{amssymb}

\usepackage{hyperref}
\usepackage{tablefootnote}
\usepackage{xspace}

\usepackage{float}
\usepackage{amsmath}
\usepackage{bm}
\usepackage{algorithmicx}
\usepackage{algorithm}
\usepackage{algpseudocode}

\usepackage{arydshln}

\usepackage{amssymb}
\usepackage{pifont}
\newcommand{\cmark}{\ding{51}}%
\usepackage{enumitem}
\usepackage{bm}

\usepackage{fontawesome5}

\usepackage{listings}
\addtocontents{toc}{\protect\setcounter{tocdepth}{-1}}

\usepackage{microtype}

\newcommand{\ours}{\textsc{AbGen}\xspace}

\newcommand{\rab}{\underline{reference ablation study}\xspace}
\newcommand{\meta}{\textsc{AbGen-Eval}\xspace}
\newcommand{\scimon}{\textsc{SciMON}\xspace}
\newcommand{\testmini}{\emph{testmini}\xspace}
\newcommand{\test}{\emph{test}\xspace}
\newcommand{\eg}{\hbox{\emph{e.g.,}}\xspace}
\newcommand{\ie}{\hbox{\emph{i.e.,}}\xspace}

\newcommand{\nmodel}{18\xspace}
\newcommand{\num}{1,500\xspace}
\newcommand{\npaper}{807\xspace}
\newcommand{\up}[1]{\textcolor{red}{(+#1)}}

\definecolor{YaleBlue}{RGB}{16, 42, 86}
\definecolor{TATABlue}{RGB}{1, 126, 199}

\newcommand{\Yale}{\hspace{.1em}^{\textcolor{YaleBlue}{\boldsymbol{Y}}}}
\newcommand{\Tata}{\hspace{.1em}^{\textcolor{TATABlue}{\boldsymbol{T}}}}

\newcommand{\huggingface}{\raisebox{-1.5pt}{\includegraphics[height=1.05em]{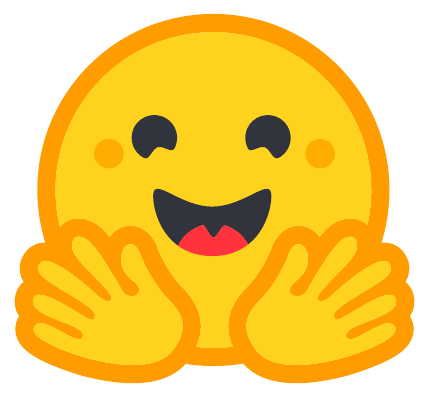}}\xspace}
\newcommand{\github}{\raisebox{-1.5pt}{\includegraphics[height=1.05em]{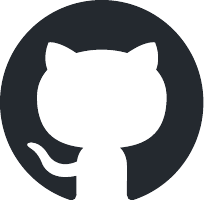}}\xspace}

\newcommand{\rqone}{How well do frontier LLMs perform in designing ablation studies?}
\newcommand{\rqtwo}{How can this research be applied in real-world scenarios to assist human researchers in designing ablation studies?}

\newcommand{\rqthree}{
How can future researchers develop more reliable and effective automated evaluation systems for complex scientific tasks?}
\title{\ours: Evaluating Large Language Models in Ablation Study \\Design and Evaluation for Scientific Research}

\author{
Yilun Zhao\thanks{~~Equal Contributions. Correspondence: Yilun Zhao (\texttt{yilun.zhao@yale.edu})}~$\Yale$ \quad Weiyuan Chen$^{*}$$\Yale$ \quad Zhijian Xu$\Yale$ \quad Manasi Patwardhan$\Tata$ \\\bf{Yixin Liu$\Yale$ \quad Chengye Wang$\Yale$ \quad Lovekesh Vig$\Tata$ \quad Arman Cohan$\Yale$} \vspace{4pt}\\
$\Yale$~Yale NLP Lab \quad $\Tata$~TCS Research
\vspace{10pt}
}

\begin{document}
\maketitle
\begin{abstract}
We introduce \textbf{\ours}, the first benchmark designed to evaluate the capabilities of LLMs in designing ablation studies for scientific research. 
\ours consists of \num expert-annotated examples derived from \npaper NLP papers. 
In this benchmark, LLMs are tasked with generating detailed ablation study designs for a specified module or process based on the given research context.
Our evaluation of leading LLMs, such as DeepSeek-R1-0528 and o4-mini, highlights a significant performance gap between these models and human experts in terms of the importance, faithfulness, and soundness of the ablation study designs.
Moreover, we demonstrate that current automated evaluation methods are not reliable for our task, as they show a significant discrepancy when compared to human assessment.
To better investigate this, we develop \textbf{\meta}, a meta-evaluation benchmark designed to assess the reliability of commonly used automated evaluation systems in measuring LLM performance on our task. We investigate various LLM-as-Judge systems on \meta, providing insights for future research on developing more effective and reliable LLM-based evaluation systems for complex scientific tasks. 

\begin{small}
\begin{center}
\begin{tabular}{cll}
\huggingface & \textbf{Data} & \href{https://huggingface.co/datasets/yale-nlp/AbGen}{\path{yale-nlp/AbGen}}\\
\github & \textbf{Code} & \href{https://github.com/yale-nlp/AbGen}{\path{yale-nlp/AbGen}}\\
\end{tabular}
\end{center} 
\end{small}
\end{abstract}

\begin{figure}[!t]
    \centering
    \includegraphics[width = \linewidth]{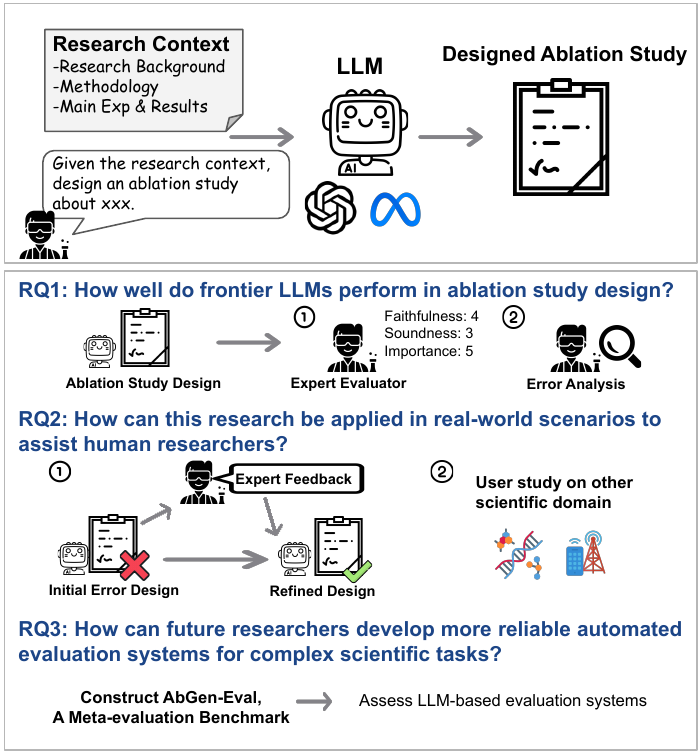}
    \caption{Overview of the research: the ablation study design task and three research questions investigated.
    }
    \label{fig:example}
\end{figure}
\begin{figure*}[!t]
\centering
\includegraphics[width=1\textwidth]{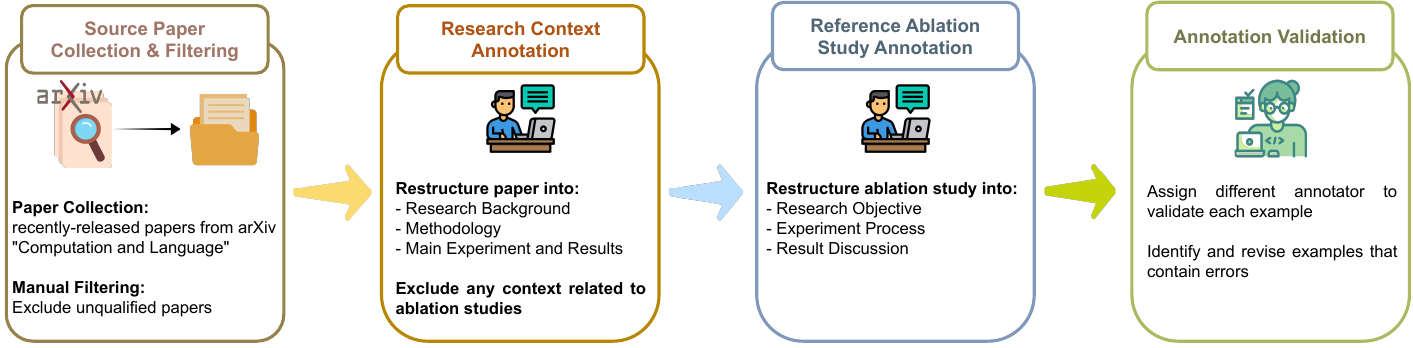}
\caption{An overview of \ours construction pipeline.
}
\label{fig:pipeline}
\end{figure*}
\section{Introduction}
In empirical scientific fields, designing experiments and selecting the appropriate experimental settings often present considerable challenges and requires significant domain expertise. Oftentimes, scientists learn about the flaws in their experimental design and missing ablations after going through a peer review process, which involves domain experts carefully evaluating a scientific work. The complexity of tasks in experimental science underscores the need for innovative approaches to support researchers in optimizing their workflows. Meanwhile, LLMs have demonstrated remarkable capabilities across a range of tasks integral to scientific processes, such as reviewing manuscripts \cite{d2024marg,du2024llms}, scientific writing \cite{ALTMAE20233,xu2024kiwi}, scientific code generation \cite{ml-bench, swe-agent}.
This raises a compelling question: \emph{Can LLMs be effectively leveraged to assist scientists in the process of experimental design?}

While addressing this question is inherently complex due to the diverse nature of scientific disciplines and difficulty of evaluation, our objective is to introduce the first comprehensive benchmark as well as an evaluation methodology to facilitate measuring progress on this task. We particularly introduce \textbf{\ours}, 
the first benchmark for evaluating LLMs in the context of designing ablation studies for scientific research.
The dataset consists of \num examples derived from \npaper scientific papers in natural language processing (NLP). 
Each example is carefully annotated and validated by NLP experts and includes a comprehensive research context along with a reference ablation study, both restructured from the original research paper. 
The research context is divided into three sections: research background, methodology, and the main experiment setup and results. 
As illustrated in \autoref{fig:example}, the LLMs are tasked with generating a detailed ablation study design for a specified module or process based on the provided research context. 

As outlined in \autoref{fig:example}, we investigate three research questions in this study.
Our main contributions are summarized below:



\begin{itemize} [leftmargin=*]
\itemsep0em
\item We propose \ours, the first benchmark designed to evaluate the capabilities of LLMs in ablation study designs for scientific research (\S\ref{sec:data}). We design a comprehensive human and automated evaluation systems for \ours (\S\ref{sec:eval}). 
\item We conduct a systematic evaluation of leading LLMs, analyzing their strengths and limitations on our new task, and providing insights for future advancements (\S\ref{sec:main-result}).
\item Our user studies reveals the potential of LLMs in ablation study design by interaction with human researchers, and highlights the adaptability of this approach to other scientific domains (\S\ref{sec:user-study}).
\item We develop the meta-evaluation benchmark, \meta, and investigate various LLM-based evaluation methods to provide insights for creating more reliable automated evaluation systems for complex scientific tasks (\S\ref{sec:meta}).

\end{itemize}
\section{\ours Benchmark}\label{sec:data}
To systematically study the capabilities and limitations of current LLMs and measuring progress in assisting scientists with the design of their experimental workflows, we introduce a new benchmark named \ours. 
The LLMs are tasked with generating detailed ablation study designs for a specified module or process based on the given research context.
We focus on scientific research within the NLP domain, as the involved expert annotators primarily have expertise in NLP (\ie each has at least one publication in a top-tier NLP or AI venue as a leading author). 
Detailed biographies of the annotators participating in the \ours annotation and LLM performance evaluation process are provided in \autoref{tab:candidate_profiles} in Appendix~\ref{app:data}.
We believe that future research could extend our benchmark construction pipeline to extend to other scientific domains.

In the following subsections, we first provide a formal definition of the \ours task and then detail each step within the benchmark construction process. We present an overview of the \ours construction pipeline in \autoref{fig:pipeline}.

\subsection{\ours Task Formulation}
We formally define the task of \ours in the context of LLMs. Specifically, given: 
\begin{itemize}[leftmargin=*]
    \itemsep0em
    \item The \textbf{research context} \bm{$C$}, which is an expert-annotated context of a specific scientific study. This context is restructured from the original paper by expert annotators, including sections of research background, methodology, and main experiment setup and results (\S\ref{sec:research-context}).
    \item The name of a specific essential module or process, denoted as \bm{$M$}, which is described in the \emph{methodology} section within research context \bm{$C$}.
\end{itemize}

\noindent The LLM is tasked with generating the design for an ablation study, \bm{$A$}, aimed at evaluating the contribution and impact of \bm{$M$} within the overall research framework:

\begin{equation}
\label{eq:formulation}
    \bm{\hat{A}} = \arg\max_A P_{\mathbf{LLM}}(\bm{A}~|~\bm{C}, \bm{M})
\end{equation}

\noindent The ablation study design should include a clear statement of the research objective, along with a detailed description of the experimental process.

\subsection{Source Paper Collection and Filtering}
\paragraph{Source Paper Collection.}
We collect scientific papers from arXiv under the ``Computation and Language'' category, targeting those first released between March 1, 2024 and August 30, 2024.
For each paper, we adopt the tool\footnote{\url{https://github.com/allenai/s2orc-doc2json}} developed by \citet{lo-etal-2020-s2orc} to extract its content.
Specifically, this tool parses LaTeX source files of papers into JSON format, extracting features including the paper title, abstract, main sections, and appendix.
We convert tables within the papers into HTML format. 
Both recent works~\cite{sui2024table, fang2024large} and our preliminary studies reveal that the evaluated LLMs can comprehend such table format effectively.
Next, we describe our approach and criteria for inclusion of the papers for annotation, as well as the details of the annotation process.

\paragraph{Research Paper Manual Filtering.} 
For each collected NLP paper, the expert annotator first determines if they are familiar with the paper's topic. If not, we randomly assign the paper to another annotator. Papers whose topics are unfamiliar to both annotators are excluded.
The annotators are then instructed to determine whether the paper qualifies for inclusion in our benchmark. Specifically, we exclude:
(1) Papers that are not focused on experimental work (\eg surveys, position papers, dissertations), as they do not involve ablation study design;
(2) Papers with fewer than two ablation studies, as these may not provide sufficient breadth of experimental evidence. 
Additionally, annotators may exclude papers they deem to be of low quality based on their expert judgment. After applying these filtering criteria, \npaper papers remain for further annotation.

\subsection{Research Context Annotation}\label{sec:research-context}
After determining that a research paper qualifies for benchmark inclusion, annotators are instructed to restructure the original paper into \underline{research context} that maintains the original meaning but exclude any content related to ablation studies. 
The \underline{research context} contains the following three sections:
(1) \textbf{Research Background}, which is restructured from the introduction and related work sections, describing the paper's motivation, research problem, and relevant prior work.
(2) \textbf{Methodology}, which is restructured from the methodology sections, This section describes the proposed method or model, including key components and innovations.
(3) \textbf{Main Experiment Setup and Results}, which is restructured from the experiment sections. This section details the primary experimental setup, including datasets, baselines, and evaluation metrics used in main experiments, as well as the main experimental results.

\subsection{Reference Ablation Study Annotation}
Annotators are then tasked with restructuring each ablation study in the research paper into a \rab.
It consists of the following three sections:
(1) \textbf{Research Objective}, a one- or two-sentence description of the research problem and the goal of the ablation study. If this statement is not explicitly provided in the original ablation study, annotators are required to infer and summarize it.
(2) \textbf{Experiment Process}, a detailed account of the experimental setup, including the experimental groups, datasets, procedures, and the evaluation tools and metrics used. Annotators are requried to ensure that the process is clearly understandable and replicable based on the provided description.
(3) \textbf{Result Discussion}, an analysis of the outcomes, where annotators summarize the key findings and their implications. It's worth noting that we do not require LLMs to generate this part, as our main focus is on evaluating their ability to design ablation studies rather than execute and analyze experiments. However, we believe these features could be valuable for future research.

\subsection{Annotation Validation}
For each annotated example, we assign an annotator to validate the annotated \underline{research context} and \rab based on the original research paper. 
They are required to identify and revise examples that contain errors.
Out of the \num annotated examples, 273 were identified as erroneous and were subsequently revised.
We conducted a final human evaluation of data quality on 100 examples. As shown in \autoref{tab:annotation_aggrement} (Appendix~\ref{app:data}), for each validation metric, over 95\% of the samples received a satisfaction rating of at least 4 out of 5. This result indicates the high quality of \ours.

\begin{table}[!t]
\centering
\small
\resizebox{\linewidth}{!}{%
\addtolength{\tabcolsep}{-0.5em}
\renewcommand{\arraystretch}{1.1}
\begin{tabular}{lr}
\toprule
\textbf{Property} & \textbf{Value} (\texttt{avg./max}) \\
\midrule
\textbf{Research Context Word Length}  & 1,847.8 / 6,253 \\
\quad Research Background  & 319.6 / 1,178 \\
\quad Methodology  & 904.4 / 4,685 \\
\quad Exp Setup \& Results  & 623.7 / 2,174 \\

\noalign{\vskip 0.5ex}\hdashline\noalign{\vskip 0.5ex}

\textbf{Ref. Ablation Study Word Length}  & 145.5 / 518 \\
\quad Research Objective  & 6.1 / 15 \\
\quad Experiment Process  & 72.5 / 264 \\
\quad Result Discussion  & 67.1 / 336 \\

\noalign{\vskip 0.5ex}\hdashline\noalign{\vskip 0.5ex}

\# NLP Research & \npaper\\
\# Ref. Ablation Study per Research  & 1.9 / 3\\
\midrule
\textbf{\ours Size} & 1,500 \\
\quad Testmini Set & 500 \\
\quad Test Set & 1,000 \\

\bottomrule
\end{tabular}
}
\caption{Data statistics of the \ours benchmark.}
\label{tab:stat}
\end{table}

\subsection{Data Statistics}
\autoref{tab:stat} illustrates the data statistics of the \ours benchmark.
We randomly split the dataset into two subsets: \testmini and \test. 
The \testmini subset contains 500 examples and is intended for both method validation and human analysis and evaluation. The \test subset comprises the remaining 1,000 examples and is
designed for standard evaluation.

\section{\ours Evaluation}\label{sec:eval}
The automated evaluation of LLM generation for tasks relevant to scientific workflows remains an unsolved problem in the community. Recent benchmark work, such as \scimon~\cite{wang2024scimon} for novel scientific direction generation and MARG~\cite{d2024marg} for peer review generation, primarily rely on human evaluation to assess LLM-based system performance. 
In our study, we also employ human evaluation by expert annotators as the \emph{primary} assessment method. 
Additionally, in Section~\ref{sec:meta}, we investigate different variants of LLM-based evaluation methods, aiming to provide insights for future work to develop automated evaluation systems for a large-scale evaluation. 

\subsection{Evaluation Criteria}
This section discusses the human and automated evaluation protocols developed for \ours evaluation.
We assess the following three dimensions for the generated ablation study design.
\begin{itemize}[leftmargin=*]
    \itemsep0em
    \item \textbf{Importance}: The generated ablation study design will provide valuable insights into understanding the role of the specified module or process within the overall methodology.
    \item \textbf{Faithfulness}: The generated ablation study design aligns perfectly with the given research context. There are no contradictions between the generated content and the main experimental setup within the provided research context.
    \item \textbf{Soundness}: The generated ablation study design is logically self-consistent without ambiguious description. The human researchers would be able to clearly understand and replicate the ablation study based on the generated context.
\end{itemize}

To determine these three dimensions, we gathered feedback from three external senior NLP researchers, all of whom serve as area chairs for the ACL Rolling Review.
Through iterative discussions, we identified these dimensions as critical for evaluating the quality and utility of generated ablation study designs. 
This feedback process also helped us in refining the assessment guidelines used for human evaluation (\S\ref{sec:eval-protocol}).
We do not evaluate the \emph{fluency} of the generated ablation study, as both recent works~\cite{d2024marg, zeng2024evaluating} and our preliminary findings find that leading LLMs consistently produce fluent text free of grammatical errors. 

\subsection{Human Evaluation Protocol}\label{sec:eval-protocol}
For human evaluation, we use Likert-scale scores ranging from 1 to 5 for each criterion (\ie importance, faithfulness, and soundness). 
Given the research context and an LLM-generated ablation study, human evaluators are asked to score the generated content for each criteria. 
Initially, the reference ablation study is not provided to the evaluator. This approach encourages evaluators to carefully review the generated content in light of the research context, reducing the likelihood of bias from comparing it to the reference. This is particularly important, as LLMs may generate ablation studies that, while reasonable, differ from the reference.
After submitting their initial scores, evaluators are then given the reference ablation study and asked to adjust their scores if they identify any aspects they may have initially overlooked.

To assess inter-annotator agreement of our human evaluation, we 
sample 40 fixed LLM-generated outputs that are separately evaluated by all four expert annotators.
They achieve inter-annotator agreement scores (\ie Cohen's Kappa) of 0.735, 0.782, and 0.710 for the criteria of importance, faithfulness, and soundness, respectively.


\subsection{Automated Evaluation} \label{sec:auto_evaluation}
While human evaluation is generally reliable, it is time-consuming and does not scale well. To address this, we also employ an LLM-as-a-judge system for automated evaluation. Specifically, we use GPT-4.1-mini as the base evaluator. For each model-generated response, the evaluator is provided with the research context and a reference ablation study. 
Evaluation is performed across four criteria (\ie importance, faithfulness, soundness, and overall quality), with the model prompted separately for each criterion to assign a score from 1 to 5. 
Prior to issuing a final score, the evaluator must generate a rationale explaining its judgment.
The full evaluation prompts used for each criterion are provided in Appendix~\ref{app:exp}.
To gain a deeper understanding of the reliability of LLM-as-Judge systems, we develop the meta-evaluation benchmark, \meta, which is detailed in Section~\ref{sec:meta}.

\section{LLMs for Ablation Study Design}\label{sec:exp-gen}
\subsection{Experiment Setup}\label{sec:setup}
\paragraph{Evaluated Systems.} We examine the performance of \nmodel frontier LLMs across two distinct categories on our benchmark: 
(1) \textbf{Proprietary LLMs}, including o4-mini~\cite{openai2025o4mini}, GPT-4o~\cite{openai2024gpt4o}, GPT-4.1~\cite{openai2024gpt4-1}, 
Gemini-2.5-Flash~\cite{geminiteam2024gemini}; 
and \textbf{Open-source LLMs}, including 
Llama-3.1-70B, Llama-3.3-70B, Llama-4-Scout-17B and Llama-4-Maverick-17B~\cite{dubey2024llama3,touvron2025llama4}, 
Mistral-Large~\cite{jiang2024mixtral},
Deepseek-V3, DeepSeek-R1-0528-Qwen3-8B, and Deepseek-R1~\cite{deepseekai2024v3,deepseekai2025r1}, 
Phi-4~\cite{microsoft2025phi4},
Gemma-3-27b-it~\cite{gemmateam2025gemma3technicalreport}
, Qwen2.5-32B, Qwen3-8B, Qwen3-32B and Qwen3-235B-A22B,~\cite{yang2024qwen2,qwen3technicalreport}.
\autoref{tab:model-info} in Appendix presents the details of these evaluated LLMs in \ours. 

\paragraph{Measuring Performance of Real Paper and Expert.}\label{app:human-performance}
To provide an informative estimate of real paper and expert-level performance on \ours, we randomly sample 20 examples from 10 papers in the \testmini set. 
We enlist two expert annotators (\ie Annotators 1 and 4, as described in \autoref{tab:candidate_profiles} in Appendix~\ref{app:data}) to individually solve these examples. 
To ensure fairness, we mix these 20$\times$2 expert-annotated data and corresponding 20 reference ablation study within the standard human evaluation process. 
The expert evaluators are not informed of the sources of these ablation study examples when evaluation. We report the evaluation results on \autoref{tab:results}.

\begin{figure}[!t]
\begin{tcolorbox}[colback=black!7.5!white, colframe=black!80!white, title=Ablation Generation Prompt, fontupper=\footnotesize, fonttitle=\footnotesize]
\texttt{[System Input]}: \vspace{2pt}\\
Given the research context, design an ablation study for the specified module or process. Begin the design with a clear statement of the research objective, followed by a detailed description of the experiment setup. Do not include the discussion of results or conclusions in the response, as the focus is solely on the experimental design. 
The response should be within 300 words. Present the response in plain text format only. \\

\texttt{[User Input]}: \vspace{2pt}\\
Research Context:\{research context\} \\
Design an ablation study about \{ablation module\} based on the research context above. \\
\end{tcolorbox}

\caption{Prompt for ablation study generation.}
\label{fig:prompt}
\end{figure}
\begin{table*}[!t]
\centering
\small
\renewcommand{\arraystretch}{1.15}
\begin{tabular}{lp{0.8cm}p{0.65cm}p{0.65cm}p{0.8cm}p{0.8cm}p{0.65cm}p{0.65cm}p{0.65cm}p{0.65cm}}
\toprule
\multirow{2}{*}{\textbf{System}}  & \multicolumn{4}{c}{\textbf{LLM-based Eval (1-5)}} & \multicolumn{4}{c}{\textbf{Human Evaluation (1-5)}}\\
\cmidrule(lr){2-5} \cmidrule(lr){6-9}
 & Import. & Faith. & Sound.& Overall  & Import. & Faith. & Sound. & \textbf{Avg.}\\
\midrule
Reference (orig) & -- & -- & --&--  & 4.70 & 4.90 & 4.70 & 4.77 \\
Human Expert & 4.82 & 4.84 & 4.33&--  & 4.65 & 4.93 & 4.83 & 4.80 \\
\midrule
DeepSeek-R1-0528 & 4.80 & \cellcolor{red!35} 4.85 & \cellcolor{red!35} 4.39 &\cellcolor{red!20}4.95& \cellcolor{red!20} 4.23 & \cellcolor{red!35} 4.00 & \cellcolor{red!35} 4.11 & \cellcolor{red!35} 4.11 \\
o4-mini & 4.80 &   4.81 & \cellcolor{red!20}  4.33&\cellcolor{red!35}4.96 & \cellcolor{red!20} 4.23 & 3.78 & 4.00 &   \cellcolor{red!20}4.00 \\
GPT-4.1 & \cellcolor{red!20} 4.82 & \cellcolor{red!20} 4.84 & 4.28&\cellcolor{red!35}4.96 &   4.12 & 3.87 &   \cellcolor{red!20}4.02 &   \cellcolor{red!20}4.00 \\
DeepSeek-V3  & 4.78 & 4.80 & 4.19&4.92 & 3.98 & 3.79 & 3.96 & 3.91 \\
Qwen3-235B-A22B & \cellcolor{red!35} 4.83 & 4.76 & 4.31&\cellcolor{red!20}4.95 & \cellcolor{red!35} 4.26 & 3.43 & 4.00 & 3.90 \\
Gemini-2.5-Flash & 4.63 & 4.52 & 4.01&4.65 & 3.89 &   \cellcolor{red!20}3.94 & 3.76 & 3.86 \\
Gemma-3-27b-it & 4.70 & 4.75 & 4.21&4.85 & 3.78 & 3.81 & 3.96 & 3.85 \\
GPT-4o &  4.81 & 4.75 & 4.15& 4.65 & 3.88 & 3.67 & 3.91 & 3.82 \\
Qwen3-32B & \cellcolor{red!20} 4.82 & 4.74 & 4.22&4.94 & 3.90 & 3.47 & 3.98 & 3.78 \\
Qwen3-8B & 4.77 & 4.69 & 4.16&4.90 & 3.86 & 3.46 & 3.89 & 3.74 \\
Mistral-Small-3.1-24B & 4.74 & 4.63 & 4.12&4.84 & 3.74 & 3.35 & 3.84 & 3.64 \\
Phi-4 & 4.74 & 4.65 & 4.12 &4.81& 3.70 & 3.34 & 3.78 & 3.61 \\
Llama-4-Maverick-17B & 4.66 & 4.64 & 4.04&4.71 & 3.46 & 3.66 & 3.68 & 3.60 \\
DeepSeek-R1-0528-Qwen3-8B & 4.69 & 4.68 & 4.12 &4.81& 3.71 & 3.18 & 3.65 & 3.51 \\
Qwen2.5-32B & 4.73 & 4.64 & 4.08 &4.80& 3.53 & 3.17 & 3.72 & 3.47 \\
Llama-4-Scout-17B & 4.71 & 4.51 & 4.04 &4.70& 3.49 & 3.22 & 3.50 & 3.40 \\
Llama-3.1-70B & 4.68 & 4.46 & 4.05 &4.70& 3.58 & 2.91 & 3.55 & 3.35 \\
Llama-3.3-70B & 4.68 & 4.45 & 4.03 &4.66& 3.27 & 3.08 & 3.49 & 3.28 \\
\bottomrule
\end{tabular}

\caption{
Human and automated evaluation results of LLMs on \ours. For automated evaluation, we use GPT-4.1-mini as the base evaluator and report scores on the \test subset. For human evaluation, we randomly sample 100 examples from the \testmini subset. Each model output is assessed by an expert evaluator. The average human score is used as the primary metric for ranking model performance in this table.
}
\label{tab:results}
\end{table*}
\paragraph{Implementation Details.}
For all the experiments, we set temperature as 1.0 and maximum output length as 1024 (as the maximum length of reference ablation study is 518 words as presented in \autoref{tab:stat}).
\autoref{fig:prompt} illustrates the default prompt used across all generation experiments.
The model is tasked with generating the design for an ablation study, based on the provided annotated research context and the specified module or process name. 
Specifically, the LLMs are required to first generate a one-sentence description of the research objectives, followed by a detailed description of the experimental setup for the ablation study.

\subsection{Results and Analysis}\label{sec:main-result}

\begin{tcolorbox}[colback=gray!10, colframe=gray!50!white, width=\linewidth, boxrule=0.5mm, arc=1mm, outer arc=1mm, left=2mm, right=2mm]
\textcolor{orange!50!yellow}{\faLightbulb} \xspace 
\textbf{RQ1:} \rqone 
\end{tcolorbox}

\noindent \autoref{tab:results} illustrates the performance of the evaluated LLMs on \ours.
The human evaluation results demonstrate that \ours poses significant challenges to current LLMs.
Even the best-performing LLM, DeepSeek-R1-0528, performs much worse than human experts. 
This gap highlights the critical need for further advancements in LLMs, especially in applying them to complex scientific tasks.
Moreover, we observe a disparity between automated evaluation systems and human assessments. 
For instance, despite receiving similar scores in LLM-based evaluations compared to o4-mini, DeepSeek-R1-0528 consistently outperforms it in every criterion according to human evaluation.
These results suggest that current automated evaluation systems may not be fully reliable for our task.
To gain a deeper understanding of the reliability of current automated evaluation systems, we develop the meta-evaluation benchmark, \meta, which is detailed in Section~\ref{sec:meta}.

\begin{table*}[!t]
\centering
\small
\begin{tabular}{p{4.1cm}p{10.7cm}}
\toprule
\textbf{Error Type} & \textbf{Explanation} \\

\midrule

Misalignment with research context 
& 
This error arises when the generated experiment process contradicts with the baseline in the research context or introduces factual errors. 
\\

\midrule

Ambiguity and Difficulty in Reproduction &
This error arises when the generated experiment process contains ambiguous steps or lacks the necessary datasets or tools, for human researchers to replicate ablation study. 
\\

\midrule
 Partial Ablation or Incomplete Experimentation &
This error arises when the generated experiment process partially addresses the ablation module, such as only ablating a sub-module, or missing experimental groups.
\\

\midrule

Insignificant Ablation Module &
This error arises when the generated research objective is focused on an insignificant ablation module in research context.
\\

\midrule

Inherent Logical Inconsistencies  &
This error arises when the generated experiment process contains inherent logical inconsistencies, such as gaps in implementation steps.
\\

\bottomrule
\end{tabular}
\caption{A summary of GPT-4o's failure cases. We provide examples for each error type in Appendix~\ref{app:err}.}
\label{tab:error}
\end{table*}

\paragraph{Error Analysis.}\label{sec:error-analysis}
We further conduct a comprehensive error analysis to better understand the capabilities and limitations of the top-performing LLMs on our task.
This error analysis is based on 100 failure cases of models from the \testmini set, where the average human evaluation scores are below 3. We identify five common error types, and provide detailed explanations for each type in \autoref{tab:error}. 
These error cases demonstrate that generating constructive ablation study designs based on research context is still challenging for LLMs.

\subsection{User Studies on Real-world Scenarios}\label{sec:user-study}

\begin{tcolorbox}[colback=gray!10, colframe=gray!50!white, width=\linewidth, boxrule=0.5mm, arc=1mm, outer arc=1mm, left=2mm, right=2mm]
\textcolor{orange!50!yellow}{\faLightbulb} \xspace 
\textbf{RQ2:} \rqtwo
\end{tcolorbox}

\begin{table}[!t]
\centering
\small
\resizebox{\linewidth}{!}{%
\renewcommand{\arraystretch}{1.1}
\addtolength{\tabcolsep}{-0.3em}
\begin{tabular}{llll}
\toprule
\textbf{User Study} & \textbf{Import.} & \textbf{Faith.} & \textbf{Sound.}\\
\midrule
\multicolumn{4}{c}{\emph{\textbf{User Study 1: LLM-Researcher Interaction}}} \\
\noalign{\vskip 2ex}

\textbf{GPT-4o} \\
Initial Failure Case & 3.9 & 2.1 & 2.0 \\
Revision with Feedback & 4.8 \up{0.9} & 4.2 \up{2.1} & 4.6 \up{2.6} \\
\noalign{\vskip 0.5ex}\hdashline\noalign{\vskip 0.5ex}
\textbf{Llama-3.1-70B} \\
Initial Failure Case & 3.7 & 1.8 & 1.7 \\
Revision with Feedback & 4.5 \up{0.8} & 3.9 \up{2.1} & 4.1 \up{2.4} \\
\midrule
\multicolumn{4}{c}{\emph{\textbf{User Study 2: Domain Generalization}}} \\
\noalign{\vskip 2ex}

\textbf{GPT-4o} \\
NLP Domain (as Main Exp) & 3.9 & 3.4 & 3.3\\
Biomedical Domain & 3.7 & 3.4 & 3.1\\
Computer Network Domain & 3.8 & 3.3 & 3.4\\
\noalign{\vskip 0.5ex}\hdashline\noalign{\vskip 0.5ex}
\textbf{Llama-3.1-70B} \\
NLP Domain (as Main Exp) & 3.3 & 2.8 & 2.8\\
Biomedical Domain & 3.0 & 2.8 & 2.9 \\
Computer Network Domain & 3.1 & 2.9 & 3.0\\

\bottomrule
\end{tabular}
}

\caption{Human evaluation result from two user studies. The findings demonstrate (1) the potential of LLMs in designing ablation studies through interaction with human researchers, and (2) the adaptability of our research across different scientific domains.}
\label{tab:user_study}
\end{table}

\noindent To investigate this research question, we design and conduct following two user studies:

\paragraph{LLM-Researcher Interaction}
While LLMs currently lag behind human experts in designing ablation studies, they still hold value as tools to assist researchers. To explore this potential, we examine scenarios where researchers interact with LLMs, providing feedback to guide the refinement of their outputs.
Specifically, we first sample 20 failure cases from \testmini set—each with an average human score below 3—from both GPT-4o and Llama-3.1-70B.
Two expert annotators are then tasked with reviewing these LLM-generated ablation study designs, identifying errors, and providing constructive feedback for improvement within a 50-word limit.
We then feed the research context, initial ablation study design, and researcher feedback back into the same LLMs, instructing them to regenerate the ablation study design.
Another expert evaluator is then assigned to assess the revised version, following the same human evaluation protocol in Section~\ref{sec:eval-protocol}.
As shown in \autoref{tab:user_study}, incorporating researcher feedback can significantly enhance LLM performance in refining their outputs.

\paragraph{Domain Generalization of Our Research.}
Our research primarily focuses on NLP domains. To explore the adaptability of our work across other scientific fields, we conducted user studies in the areas of biomedical sciences and computer networks.
Specifically, we engage two experts—one in computer networking and one in biomedical research—to provide five research papers from their respective fields that were first published after May 1, 2024, and with which they are familiar.
Following the same procedure as \ours annotation, they annotate the research context and reference ablation studies from five corresponding papers, resulting in a total of 27 examples over ten papers. 
We then provide them with LLM-generated ablation study designs and ask them to strictly follow our human assessment guidelines to evaluate the LLM outputs.
As shown in \autoref{tab:user_study}, the human evaluation scores for GPT-4o and Llama-3.1-70B are consistent with the results observed in the NLP domain experiments. We believe that future work could extend our research framework to other scientific domains.

\section{Investigating Automated Evaluation for Ablation Study Design}\label{sec:meta}
\begin{tcolorbox}[colback=gray!10, colframe=gray!50!white, width=\linewidth, boxrule=0.5mm, arc=1mm, outer arc=1mm, left=2mm, right=2mm]
\textcolor{orange!50!yellow}{\faLightbulb} \xspace 
\textbf{RQ3:} \rqthree
\end{tcolorbox}

\noindent As discussed in Section~\ref{sec:main-result}, we observe a significant discrepancy between automated and human evaluation results when assessing LLM performance on \ours.
To investigate this issue further, we conduct a systematic meta-evaluation of commonly used automated evaluation systems.

\subsection{\meta Benchmark}
We construct the meta-evaluation benchmark, \meta, based on the human assessments results collected in Section~\ref{sec:exp-gen}. 
\meta comprises $\nmodel$ LLM outputs $\times$ $100$ human assessments $= 1,800$ examples.
Each example includes an LLM-generated ablation study design and three human scores assessing the study's importance, faithfulness, and soundness, respectively (detailed in \S\ref{sec:eval-protocol}).
In line with previous meta-evaluation studies~\cite{fabbri-etal-2021-summeval, chen-etal-2021-factuality-checkers, liu-etal-2024-benchmarking}, in \meta, the human evaluation results on the system-generated ablation study is considered the gold standard. 

The performance of automated evaluation systems is measured by the \textbf{system-level} and \textbf{instance-level} correlation between scores of human evaluation and automated evaluation systems.
Specifically, given $n$ input scientific papers and $m$ ablation study generation systems, the human evaluation and an automatic metric result in two $n$-row, $m$-column score matrices $H$, $M$ respectively. The \textit{system}-level correlation is calculated on the aggregated system scores:
\begin{equation}
\label{eq:sys_corr}
    r_{\mathrm{sys}}(H, M) = \mathcal{C}(\bar{H}, \bar{M}), 
\end{equation}
where $\bar{H}$ and $\bar{M}$ contain $m$ entries which are the average system scores across $n$ data samples (e.g., $\bar{H}_0 = \sum_i H_{i,0} / n$), and $\mathcal{C}$ is a function calculating a correlation coefficient (e.g., the Pearson's correlation coefficient).
In contrast, the \textit{instance}-level correlation is an average of sample-wise correlations:
\begin{equation}
\label{eq:summ_corr}
    r_{\mathrm{sum}}(H, M) = \frac{\sum_i \mathcal{C}(H_i, M_i)}{n}, 
\end{equation}
where $H_i$, $M_i$ are the evaluation results on the $i$-th data sample.

\begin{table}[!t]
\centering
\small
\resizebox{\linewidth}{!}{%
\renewcommand{\arraystretch}{1.1}
\addtolength{\tabcolsep}{-0.3em}
\begin{tabular}{lrrrr}
\toprule
\textbf{Evaluator LLM} & \textbf{Import.} & \textbf{Faith.} & \textbf{Sound.} & \textbf{Overall} \\
\midrule
Gemini-2.5-Flash & \textbf{0.391} & \textbf{0.482} & \textbf{0.378} & \textbf{0.307} \\
    Qwen3-32B                & 0.305 & 0.405 & \underline{0.299} & \underline{0.248} \\
    GPT-4.1                  & 0.238 & \underline{0.445} & 0.298 & 0.246 \\
    DeepSeek-R1-0528         & \underline{0.352} & 0.234 & 0.070 & 0.245 \\
    Qwen3-8B                 & 0.318 & 0.308 & 0.298 & 0.237 \\
    QwQ-32B                  & 0.232 & 0.338 & 0.284 & 0.225 \\
    GPT-4.1-mini             & 0.164 & 0.329 & 0.193 & 0.194 \\
    GPT-4o                   & 0.151 & 0.249 & 0.139 & 0.179 \\
    Llama-3.3-70B            & 0.102 & 0.268 & 0.239 & 0.170 \\
    Qwen2.5-32B              & 0.109 & 0.234 & 0.173 & 0.144 \\
    DS-R1-0528-Qwen3-8B      & 0.232 & 0.265 & 0.253 & 0.124 \\
    Llama-4-Maverick         & 0.158 & 0.038 & 0.136 & 0.122 \\
    Llama-3.1-70B            & 0.071 & 0.100 & -0.020 & 0.100 \\
    Llama-4-Scout            & 0.167 & 0.026 & 0.105 & 0.083 \\
\bottomrule
\end{tabular}
}
\caption{Instance-level Pearson correlations between pointwise evaluations from various LLM-based evaluators and human judgments across four criteria: \emph{importance}, \emph{faithfulness}, \emph{soundness}, and \emph{overall}. 
The \emph{overall} score is not directly rated by humans, but computed as the average of the other three aspect scores. Evaluation prompts used in the LLM-based pairwise evaluations for each aspect are provided in Appendix~\ref{app:exp}. The system-level correlations are presented in \autoref{tab:system_level} in Appendix.}

\label{tab:instance_level}
\end{table}

\subsection{Experiments}\label{sec:meta-result}
For the LLM-based evaluation systems, we developed multiple variants to investigate how different factors influence their effectiveness. These factors include: the choice of base LLMs, ranging from open-source to proprietary models; and whether evaluation is based on specific criteria or overall scores.
As illustrated in \autoref{tab:instance_level} and \autoref{tab:system_level} in Appendix, the current automated evaluation systems show relatively low correlations, indicating that they are not reliable for assessing generated ablation study designs.
We believe future research could build on \meta dataset to develop more advanced and robust LLM-based evaluation methods for scientific tasks.
\section{Related Work}
LLMs have been employed for different scientific tasks for enhancing researchers' scientific workflows, such as 
conducting literature reviews~\cite{wang2024surveyagent, agarwal2024litllm}, 
question answering over scientific papers~\cite{dasigi-etal-2021-dataset,Saikh2022ScienceQAAN, pmlr-v202-lee23n,li-etal-2024-m3sciqa, wang2025sciver, zhao2025multimodalfoundationmodelsunderstand}, 
research hypothesis generation~\cite{wang2024scimon, zhou2024hypothesis, si2025can},
scientific paper writing~\cite{xu2024kiwi, lu2024aiscientistfullyautomated}, 
and peer-review and meta-review generation~\cite{d2024marg, tan2024peer,Wu2022IncorporatingPR,Zhou2024IsLA, xu2025llmsidentifycriticallimitations},
However, the potential of LLMs to effectively assist scientists in the experimental design process remains largely open research questions~\cite{li2024mlrcopilot, lou2025aaar, chen2025mlrbench}. Additionally, the challenge of developing effective and reliable automated evaluation systems for complex scientific tasks is underexplored~\cite{zhao2025sciarena}. Our work bridges these gaps by introducing standard benchmarks for evaluating both ablation study design and evaluation.
\section{Conclusion}

This paper introduces \ours, the first benchmark designed to evaluate LLMs in generating ablation studies for scientific research. Through a comprehensive assessment, we highlight both the strengths and limitations of leading LLMs on \ours, providing valuable insights for future advancements. 
Our findings offer practical guidance on how to apply this research in real-world scenarios, ultimately aiding human researchers. 
Additionally, 
we identify a discrepancy between automated evaluations and human assessments in our task. To investigate this, 
we also develop a meta-evaluation benchmark, providing insights into developing more reliable automated evaluation for complex scientific tasks.

\section*{Acknowledgments}
This project is supported by Tata Sons Private Limited, Tata Consultancy Services Limited, and Titan. 
We are grateful to Nvidia Academic Grant Program for providing computing resources.
\section*{Limitations and Future Work}
This study does not explore advanced prompting techniques~\cite{yao2023tree, wang2024scimon} or LLM-Agent-based methods~\cite{d2024marg, majumder2024data}. 
Our focus is on assessing the fundamental capabilities of leading LLMs in ablation study design. 
The goal is to provide insights into their strengths and limitations, laying the groundwork for future advancements. 
We encourage researchers to build upon our benchmark and findings to develop more advanced approaches for this task.
Second, as shown in our results on \meta, the reported automated evaluation scores are not yet perfect. To support further research, we will make all model outputs from Section~\ref{sec:exp-gen} publicly available. This will enable other researchers to conduct different automated evaluations and ensure consistent rankings by re-running their assessments on our model outputs.
Additionally, our human evaluation protocol is designed to minimize the need for repeated human evaluations by future researchers. By strictly adhering to our assessment guidelines, researchers can reliably assess and compare their methods with existing approaches in an independent and consistent manner.
Lastly, we only explore the LLMs' abilities on designing ablation studies. In real-world scenarios, how can LLM execute the designed ablation studies would be an interesting topic and we encourage future work to explore~\cite{chen2025ai4researchsurveyartificialintelligence}.

\bibliography{anthology,custom, llm}

\appendix


\newpage
\onecolumn
\section{Appendix}
\subsection{\ours Benchmark}\label{app:data}
\begin{table}[h]
\centering
\small
\begin{tabular}{lccc}
\toprule
\textbf{Annotation Quality}    & \textbf{\%S $\geq$ 4}\\
\midrule
\textbf{Research Context} \\
\quad Correctly structured  &  99.0 \\
\quad Excluding ablation-relevant content & 96.5 \\
\noalign{\vskip 0.5ex}\hdashline\noalign{\vskip 0.5ex}
\textbf{Reference Ablation Study} \\
\quad Correctly structured & 98.5 \\
\quad Non-overlapping & 96.0 \\
\quad Justifiable within research context & 97.5 \\
\bottomrule
\end{tabular}

\caption{Human evaluation over 200 samples of \ours. Three internal evaluators were asked to rate the samples on a scale of 1 to 5 individually. We report percent of samples that have an average score $\geq$ 4 to indicate the annotation quality of \ours.}
\label{tab:annotation_aggrement}
\end{table}
\begin{table*}[h]
\centering
\small
\begin{tabular}{llcccc}
\toprule
\textbf{ID} & \textbf{\# NLP/AI Publication} & \textbf{Data Annotation} & \textbf{Data Validation} & \textbf{Human Evaluation} & \textbf{Human Performance}\\
\midrule
1 & > 10 & \cmark & \cmark &  & \cmark\\

2 & > 10 & & & \cmark \\

3 & > 10 & & & \cmark \\

4 & 5-10 & \cmark & \cmark & & \cmark\\

5 & 1-5 & \cmark &  & \cmark \\

6 & 1-5 & \cmark & \cmark & \cmark \\
\bottomrule
\end{tabular}
\caption{Details of annotators involved in dataset construction and LLM performance evaluation. \ours is annotated by experts in NLP domains, ensuring both the accuracy of the benchmark and the reliability of the human evaluation.}
\label{tab:candidate_profiles}
\end{table*}

\section{Experiment Setup}\label{app:exp}
\begin{figure}[h]
\begin{tcolorbox}[colback=black!7.5!white, colframe=black!80!white, title=User Study Prompt, fontupper=\footnotesize, fonttitle=\footnotesize]
\texttt{[System Input]}: \vspace{2pt}\\
Revise or rewrite the initial generation based on research context and user feedback. \\

\texttt{[User Input]}: \vspace{2pt}\\
Research context: \{research context\} \\
Initial generation: \{initial generation\} \\
User feedback: \{user feedback\} \\
\newline
Redesign an ablation study about the \{ablation module\}, according to user feedback … 
\end{tcolorbox}

\caption{Prompt for LLM-researcher interaction.}
\label{fig:user_prompt}
\end{figure}

\begin{table*}[h]
\centering
\small
\begin{tabular}{llllcrrc}
\toprule
\textbf{Organization} & \textbf{Model} & \textbf{Release} & \textbf{Version} & \textbf{\begin{tabular}[c]{@{}c@{}}Context\\Window\end{tabular}} &  \\
\midrule
\multicolumn{8}{c}{\emph{\textbf{Proprietary 
 Models}}} \\
 \midrule
\multirow{2}{*}{OpenAI} &o4-mini &2025-4 & \texttt{o4-mini-2025-04-16} & -- \\
& GPT-4.1 &2025-4 & \texttt{gpt-4.1-2025-04-14} & --\\
& GPT-4o &  2024-8  &  \texttt{gpt-4o-2024-08-06}  &-- \\

\noalign{\vskip 0.5ex}\hdashline\noalign{\vskip 0.5ex}

\multirow{1}{*}{Google} & Gemini-2.5-Flash & 2024-5 & \texttt{gemini-2.5-flash-preview-05-20} &--  \\

\midrule
\multicolumn{8}{c}{\emph{\textbf{Open-source Multimodal Foundation Models}}} \\
 \midrule

\multirow{1}{*}{Mistral AI} 
& Mistral-Small-3.1 & 2025-3 & \texttt{Mistral-Small-3.1-24B} &128k \\

\noalign{\vskip 0.5ex}\hdashline\noalign{\vskip 0.5ex}

\multirow{1}{*}{Microsoft} 
&Phi-4&2025-3&\texttt{Phi-4}&16k\\
\noalign{\vskip 0.5ex}\hdashline\noalign{\vskip 0.5ex}
\multirow{1}{*}{Google} 
&Gemma-3-27b-it&2025-3&\texttt{gemma-3-27b-it}&16k\\

\noalign{\vskip 0.5ex}\hdashline\noalign{\vskip 0.5ex}

\multirow{3}{*}{DeepSeek} 
&DeepSeekV3 & 2024-12 & \texttt{DeepSeekV3} &160k \\
& DeepSeekR1 & 2025-5 & \texttt{DeepSeek-R1-0528} &160k\\ 

& DeepSeek-R1-0528-Qwen3-8B, & 2025-5 & \texttt{DeepSeek-R1-0528-Qwen3-8B} &160k\\ 
\noalign{\vskip 0.5ex}\hdashline\noalign{\vskip 0.5ex}


\multirow{4}{*}{Alibaba} 
& Qwen2.5-32B & 2025-1 & \texttt{Qwen2.5-32B-Instruct} &32k \\
& Qwen3-8B & 2025-5 & \texttt{Qwen3-8B} &40k\\ 

& Qwen3-32B & 2025-5 & \texttt{Qwen3-32B} &40k\\ 

& Qwen3-235BA22B & 2025-5 & \texttt{Qwen3-235B-A22B} &32k \\
\noalign{\vskip 0.5ex}\hdashline\noalign{\vskip 0.5ex}


\multirow{4}{*}{Meta} 
& Llama-3.1-70B & 2024-6 & \texttt{Llama-3.1-70B-Instruct} &32k \\
& Llama-3.3-70B& 2025-5 & \texttt{Llama-3.3-70B-Instruct} &32k\\ 

& Llama-4-Scout-17B& 2025-5 & \texttt{Llama-4-Scout-17B-Instruct} &32k\\ 

& Llama-4-Maverick-17B& 2025-5 & \texttt{Llama-4-Maverick-17B-Instruct} &32k \\

\bottomrule
\end{tabular}

\caption{Details of the organization, release time, maximum context length, and model source (\ie url for proprietary models and Huggingface model name for open-source models) for the LLMs evaluated in \ours.}
\label{tab:model-info}
\end{table*}




\clearpage
\section{Experiments}\label{app:meta-exp}\label{appendix:corr}

\subsection{Meta Evaluation Results}
\begin{table}[H]
\centering
\small
\renewcommand{\arraystretch}{1.1}
\addtolength{\tabcolsep}{-0.3em}
\begin{tabular}{lrrrr}
\toprule
\textbf{Evaluator LLM} & \textbf{Import.} & \textbf{Faith.} & \textbf{Sound.} & \textbf{Overall} \\
\midrule
QwQ-32B                     & \textbf{0.856} & 0.682 & 0.858 & \textbf{0.877} \\
Qwen3-32B                   & 0.741 & \textbf{0.779} & \textbf{0.884} & \underline{0.864} \\
Qwen3-8B                    & \underline{0.796} & 0.682 & 0.818 & 0.847 \\
Gemini-2.5-Flash-Preview    & 0.590 & 0.748 & 0.849 & 0.775 \\
GPT-4o                      & 0.473 & 0.607 & 0.767 & 0.726 \\
GPT-4.1-mini                & 0.562 & 0.523 & 0.828 & 0.713 \\
Qwen2.5-32B                 & 0.342 & 0.673 & 0.687 & 0.673 \\
DS-R1-0528-Qwen3-8B         & 0.674 & \underline{0.757} & 0.862 & 0.660 \\
GPT-4.1                     & 0.606 & 0.678 & \underline{0.864} & 0.647 \\
Llama-4-Maverick            & 0.584 & 0.241 & 0.622 & 0.523 \\
Llama-3.3-70B               & 0.463 & 0.404 & 0.841 & 0.516 \\
Llama-3.1-70B               & 0.264 & 0.409 & 0.266 & 0.436 \\
Llama-4-Scout               & 0.620 & 0.327 & 0.409 & 0.421 \\
DeepSeek-R1-0528            & 0.752 & 0.691 & 0.181 & 0.407 \\
\bottomrule
\end{tabular}
\caption{System-level Kendall correlations between pointwise evaluations from various LLM-based evaluators and human judgments across four criteria: \emph{importance}, \emph{faithfulness}, \emph{soundness}, and \emph{overall}. 
The \emph{overall} score is not directly rated by humans, but computed as the average of the other three aspect scores.}

\label{tab:system_level}
\end{table}

\newpage
\section{Error Analysis}\label{app:err}
\subsection{Misalignment with Research Context}

\begin{figure}[htbp]
\centering
\includegraphics[width=0.75\textwidth]{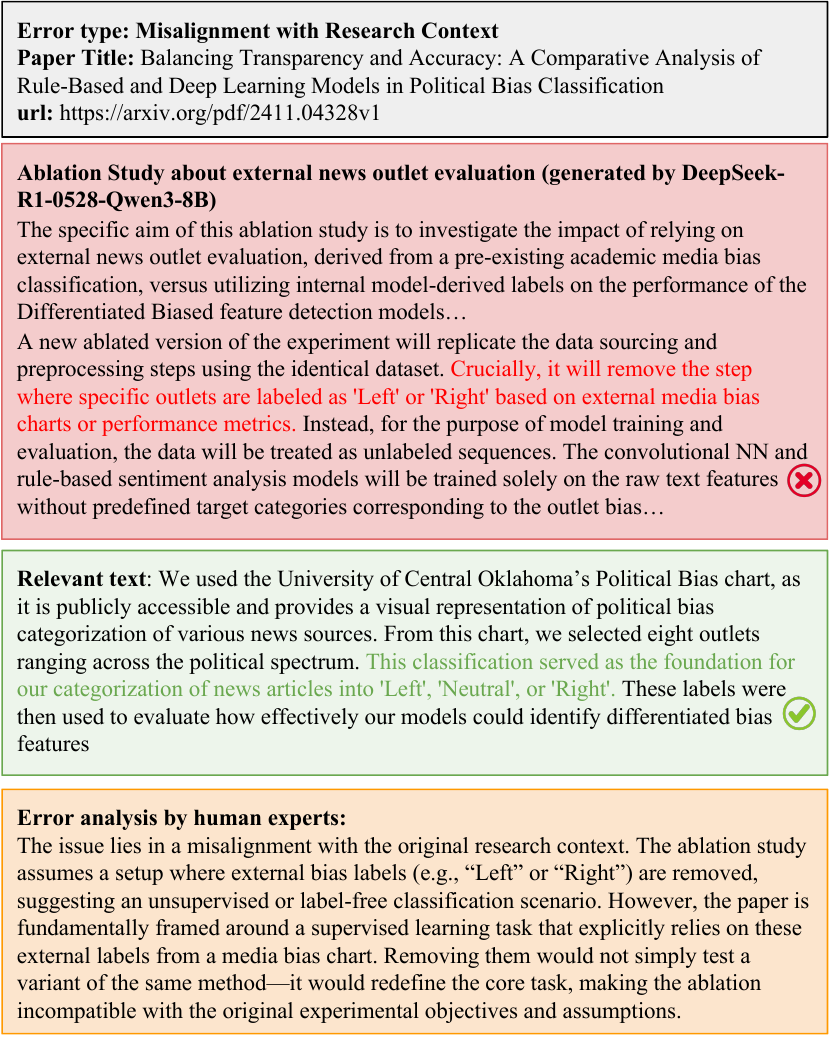}
\caption{A Failure Example of Misalignment with Research Context}
\label{fig:e1}
\end{figure}

\newpage
\subsection{Ambiguity and Difficulty in Reproduction}
\begin{figure}[htbp]
\centering
\includegraphics[width=0.75\textwidth]{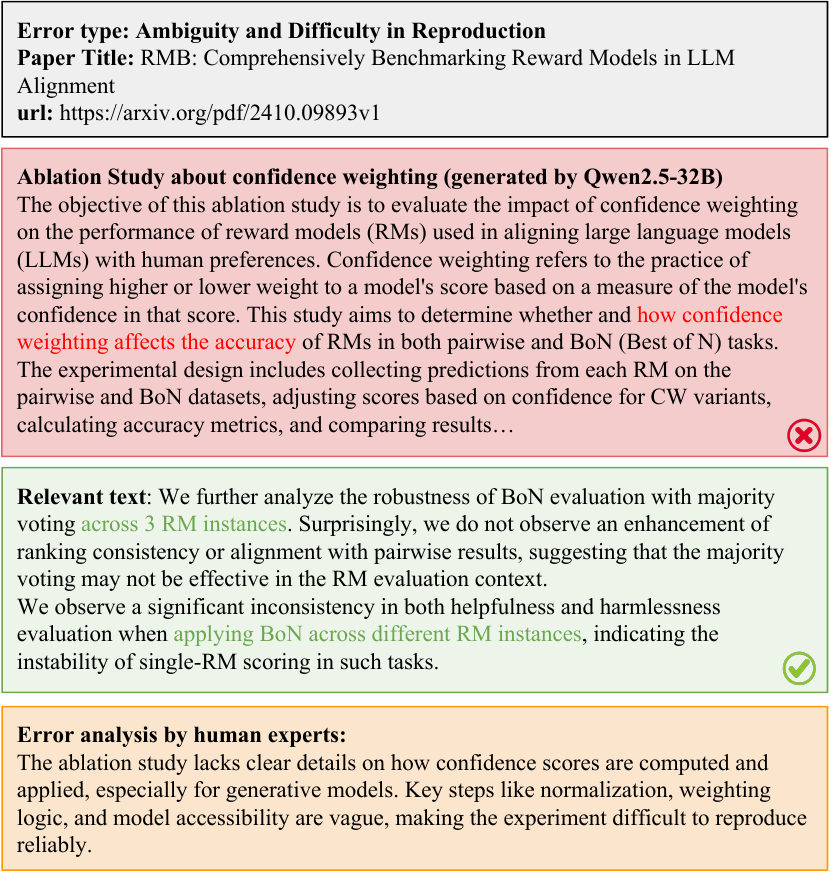}
\caption{A Failure Example of Ambiguity and Difficulty in Reproduction}
\label{fig:e2}
\end{figure}

\newpage
\subsection{Partial Ablation or Incomplete Experimentation}
\begin{figure}[htbp]
\centering
\includegraphics[width=0.75\textwidth]{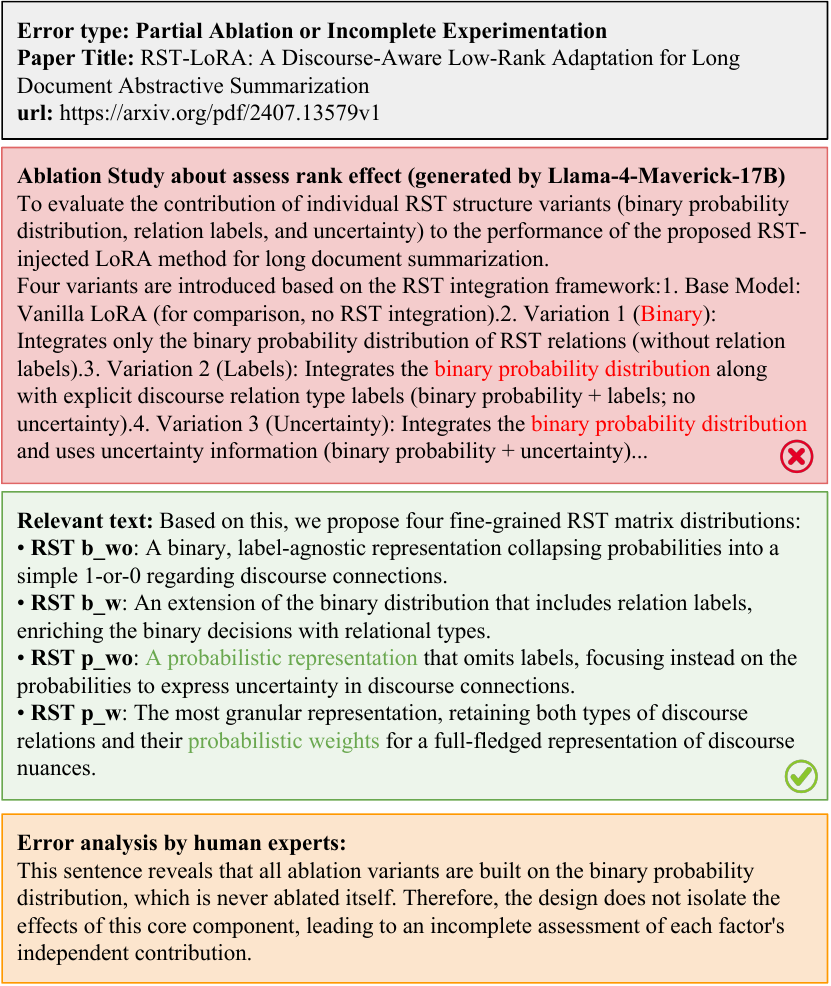}
\caption{A Failure Example of Partial Ablation or Incomplete Experimentation}
\label{fig:e3}
\end{figure}

\newpage
\subsection{Insignificant Ablation Module}
\begin{figure}[htbp]
\centering
\includegraphics[width=0.75\textwidth]{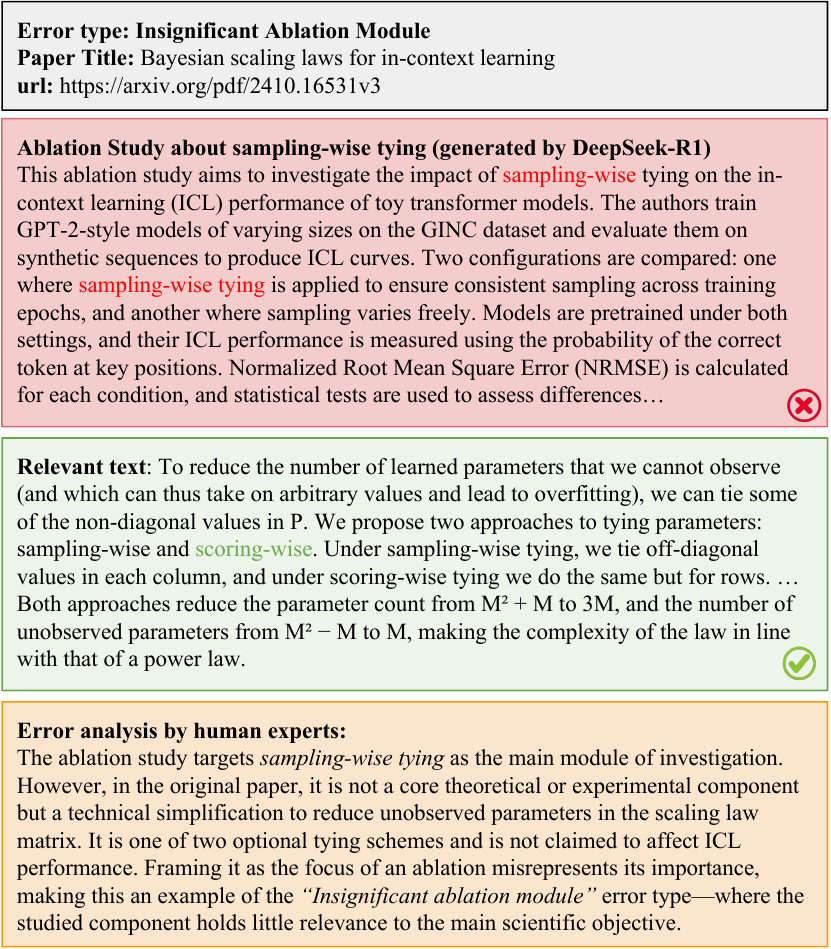}
\caption{A Failure Example of Insignificant Ablation Module}
\label{fig:e4}
\end{figure}

\newpage
\subsection{Inherent Logical Inconsistencies}
\begin{figure}[htbp]
\centering
\includegraphics[width=0.75\textwidth]{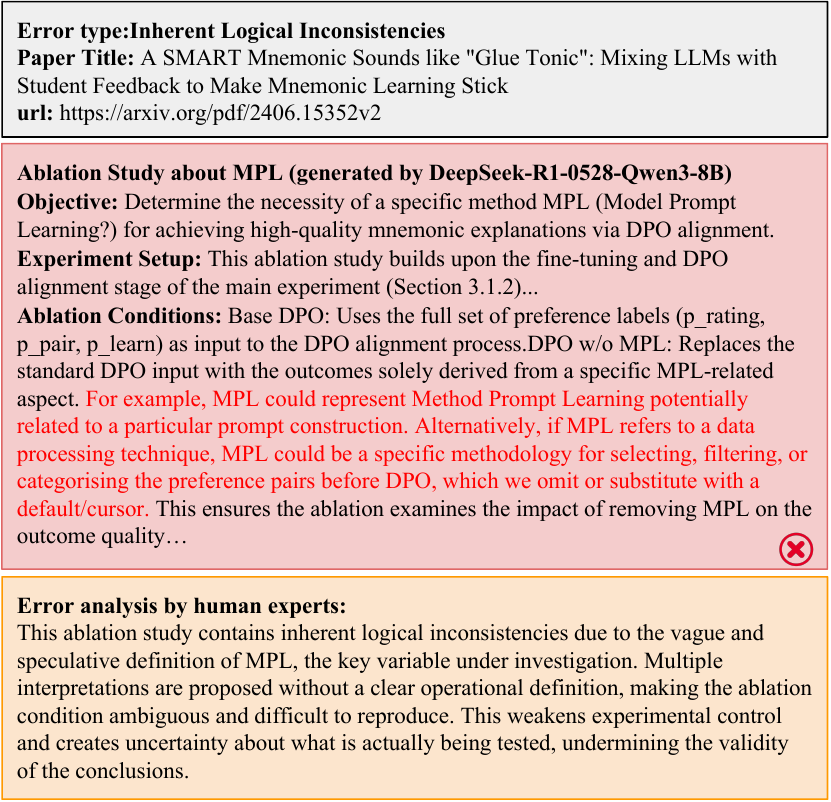}
\caption{A Failure Example of Inherent Logical Inconsistencies}
\label{fig:e5}
\end{figure}

\end{document}